\documentclass[letterpaper,journal,final]{IEEEtran}
\usepackage{amsmath,amsfonts}
\usepackage{algorithmic}
\usepackage{algorithm}
\usepackage{array}
\usepackage{textcomp}
\usepackage{stfloats}
\usepackage{url}
\usepackage{verbatim}
\usepackage{graphicx}
\usepackage[hidelinks]{hyperref}
\usepackage{orcidlink}
\usepackage[english]{babel}
\usepackage{ragged2e}
\usepackage{mathtools}
\usepackage{tikz}
\usetikzlibrary{arrows.meta,arrows}
\usepackage{tikz-3dplot}
\usepackage{amsmath}
\usepackage{amssymb}
\usepackage{ifthen}

\usepackage{bm}
\renewcommand{\vec}{\bm}
\usepackage{siunitx}
\usepackage{chemformula}
\usepackage{multirow}

\usepackage{caption}
\usepackage{subcaption}
\usepackage[nolist,nohyperlinks,printonlyused]{acronym}

\newcommand{\makeEmptyGrid}{%
\begin{tikzpicture}
    \def \scale {0.1};
    \foreach \x in {0,...,7} {
        \foreach \y in {0,...,3} {
            \draw (\scale*\x, \scale*\y) rectangle (\scale*\x+\scale, \scale*\y+\scale);
        }
    }
\end{tikzpicture}
}

\newcommand{\makeGrid}[1]{%
\begin{tikzpicture}
    \def \scale {0.1};
    \foreach \x in {0,...,7} {
        \foreach \y in {0,...,3} {
            \draw (\scale*\x, \scale*\y) rectangle (\scale*\x+\scale, \scale*\y+\scale);
        }
    }

    \foreach \x/\y in {
        #1
    }{
        \fill (\x*\scale-0.5*\scale, \y*\scale-0.5*\scale) circle (0.4*\scale);
    }
\end{tikzpicture}
}

\begin{document}

\begin{acronym}
\acro{IMU}{Inertial Measurement Unit}
\acro{CAD}{Computer Aided Design}
\acro{ESC}{Electronic Speed Controller}
\acro{CoG}{Centre of Gravity}
\end{acronym}

\title{Rapid and Inexpensive Inertia Tensor Estimation from a Single Object Throw}
\author{Till M. Blaha$^*$\,\orcidlink{0009-0006-0881-1002}, Mike M. Kuijper$^*$\,\orcidlink{0000-0002-7933-7006}, Radu Pop\,\orcidlink{0009-0000-0611-2087}, Ewoud J.J. Smeur\,\orcidlink{0000-0002-0060-6526}
\thanks{© 2025 IEEE.  Personal use of this material is permitted.  Permission from IEEE must be obtained for all other uses, in any current or future media, including reprinting/republishing this material for advertising or promotional purposes, creating new collective works, for resale or redistribution to servers or lists, or reuse of any copyrighted component of this work in other works.}
\thanks{Manuscript received: Month, Day, Year; Revised Month, Day, Year; Accepted Month, Day, Year.}
\thanks{All authors are with the Faculty of Aerospace Engineering, Delft University of Technology, The Netherlands.} \thanks{e-mail: {\tt\footnotesize \{t.m.blaha|e.j.j.smeur\}@tudelft.nl},\\ {\tt\footnotesize \{m.m.kuijper|r.pop\}@student.tudelft.nl}}%
\thanks{Till Blaha and Mike Kuijper are co-first authors. Till Blaha is corresponding.}%
\thanks{Digital Object Identifier (DOI): see top of this page.}}

\markboth{IEEE Robotics and Automation Letters,~Vol.~XX, No.~XX, Date}
{Blaha \MakeLowercase{\textit{et al.}}: Inertia estimation}

\IEEEpubid{0000--0000/00\$00.00~\copyright~2024 IEEE}

\maketitle

\begin{abstract}
The inertia tensor is an important parameter in many engineering fields, but measuring it can be cumbersome and involve multiple experiments or accurate and expensive equipment.
We propose a method to measure the moment of inertia tensor of a rigid body from a single spinning throw by attaching a small and inexpensive stand-alone measurement device consisting of a gyroscope, accelerometer, and a reaction wheel.
The method includes a compensation for the increase of moment of inertia due to adding the measurement device to the body, and additionally obtains the location of the centre of gravity of the body as an intermediate result.
Experiments performed with known rigid bodies show that the mean accuracy is around 2\%. Monte Carlo simulations reveal invariance to direction of spin and positioning of the measurement device, but show some sensitivity to noise.
\end{abstract}


\section{Introduction}

\IEEEPARstart{I}{n} numerous applications, the mass properties of a rigid body are essential quantities required for computing its dynamic behaviour.
This applies to various fields, from physics to engineering and robotics.
For example, it is required for accurate simulation and control of satellites, robots, and (flying) vehicles.
While the mass of an object can often be easily and directly measured, the inertia tensor governing the rotational motion cannot.

When the mass distribution of an object is known, e.g., from a \ac{CAD} programme, then its inertia can be calculated in a straightforward fashion.
However, the number of parts that make up a rigid body can be very high, and not every small part may be accounted for. 
Moreover, manufacturing imperfections may lead to deviations from the predicted inertia tensor in practice.
Finally, 3D models with accurate densities may not be available for all objects of interest.
In these cases, the inertia of an object has to be determined experimentally.

The most straightforward method is to measure the inertia tensor by restricting the object to rotate around one axis and applying a known torque on this axis.
From the angular acceleration, the inertia about this axis can be determined \cite{schedlinskiSurveyCurrentInertia2001}.
For this method, the object needs to be accurately mounted on a turntable, potentially requiring an adapter.
The accuracy of the method will largely depend on the accuracy with which the applied torque is known, necessitating sophisticated hardware.
Measurements about multiple axes will have to be performed in order to get the full inertia tensor, and even more measurements if the location of the centre of gravity is unknown as well.

Alternatively, there are pendulum methods, which do not require sophisticated hardware, aside from a means to measure the oscillation period (e.g., optically, or an onboard gyroscope).
With the gravitational pendulum method, the object is attached to an axis in the horizontal plane and is let free to oscillate.
The most accurate results are obtained if the objects' centre of gravity is close to the radius of gyration about that axis \cite{Dowling2006TheInertia}.
With bifilar pendulum \cite{fukami2019, setati2022} or multi-filar pendulum methods \cite{Hou2009AAssembly}, the axis of rotation is the vertical axis.
The bifilar pendulum method is a popular low-cost method for vehicles that have gyroscopes onboard.
Although these pendulum methods can produce accurate results, they require multiple experiments to obtain the full inertia tensor, and the workload is high due to the requirement of repeatedly suspending the object with wires.
In some cases, compensation for aerodynamic effects may be needed \cite{lehmkuhlerMethodsAccurateMeasurements2016}.

\begin{figure}
    \centering
    \includegraphics[trim={0.7cm, 0cm, 0.7cm, 0.7cm}, clip, width=\columnwidth]{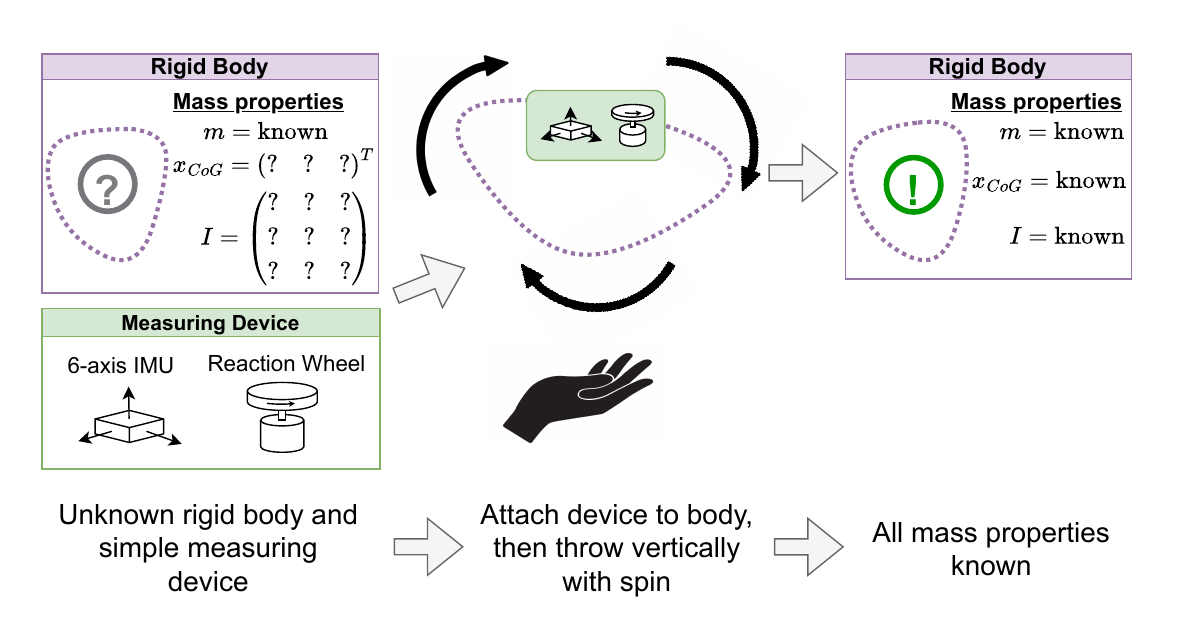}
    \caption{We propose a simple measuring device and procedure to determine inertial properties.}
    \label{fig:graphical_abstract}
    \vspace{-2mm}
\end{figure}

\IEEEpubidadjcol

For fixed-base multi-link robots, a common technique is to use least squares estimation based on the dynamic equation and data from a sufficiently excited robot with precisely known joint torques \cite{leboutetInertialParameterIdentification2021}.
It is often difficult to provide enough excitation to identify all inertial parameters independently.

The contribution of this paper is a method to apply these dynamic identification techniques to accurately estimate the mass moments of inertia and centre of gravity of a small body in seconds, by only making use of inertial measurements during a single vertical spinning throw of the object.
These measurements are obtained from a low-cost device mounted to the object containing a consumer-grade \ac{IMU}, measuring angular velocities and linear accelerations, together with a reaction wheel capable of delivering a torque. We verify that with enough spin, torque around a single axis is enough to identify all inertial parameters. \footnote{Data and analysis code: \url{https://github.com/tudelft/inertiaEstimation}}

An approach from the literature that is conceptually close is \cite{gabicciniStateInertialParameter2019}, where inertia parameters are estimated from frictional impacts of free-falling objects.
However, this problem is more complex than the problem that we investigate, leading to a much more complex method and setup.

\section{Methodology}\label{sec:method}
In the absence of external forces, the rotational behaviour of a rigid body is completely determined by its inertia tensor.
This tensor can be represented by
\begin{equation}
    \left(I_{ij}\right) = \int_{\Omega} \left(||\vec{\rho}||^2\delta_{ij} - \rho_i\rho_j\right)~dm,
\end{equation}
for infinitesimal masses positioned at $\vec{\rho}$ from the centre of mass in region $\Omega$, where $\delta_{ij}$ is the Kronecker delta. This is a linear operation, and can thus be summed over different bodies.

Euler's rotation equation states that
\begin{equation}
    I\dot{\vec{\omega}} + \vec{\omega} \times \left(I\vec{\omega}\right) = \vec{M},
\end{equation}
where $I$ is the body's inertia tensor with respect to its centre of mass, $\vec{\omega}$ is the body's angular rotation vector, $\dot{\vec{\omega}}$ is the body's angular acceleration, and $\vec{M}$ is an external torque. In the absence of an external torque ($\vec{M} = \vec{0}$), this equation is homogeneous, which makes rotation data without external torques ineffective for estimating $I$.\footnote{Relative moments of inertia can be found, but not their absolute values.}

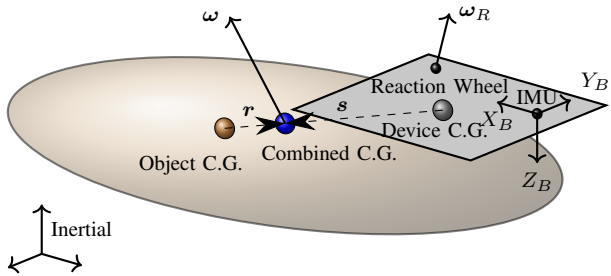
\begin{figure}[!h]
    \centering
    \tikzset{
  font={\fontsize{8pt}{12}\selectfont}}
\tdplotsetmaincoords{70}{130}
\begin{tikzpicture}[tdplot_main_coords, scale=0.7]

\definecolor{potato}{RGB}{210, 180, 140}
\definecolor{plateCol}{RGB}{200, 200, 200}

\coordinate (inertial) at (5,-2,0.5);
\draw[thick,->] (inertial) -- ++(1,0,0) node[anchor=south] {};
\draw[thick,->] (inertial) -- ++(0,1,0) node[anchor=west] {};
\draw[thick,->] (inertial) -- ++(0,0,1) node[anchor=east] {};
\node at ($ (inertial) + (0,0,0.5) $) [anchor=west] {Inertial};

\shade[ball color=potato, opacity=0.5, draw=black, thick] (-1,-1,1.25) ellipse (4.0 and 6.0);

\coordinate (potato) at (-1,-2.5,1.25);
\node at ($ (potato) + (-0.2,-1,-0.65)$) [anchor=north] {Object C.G.};

\coordinate (plate) at (0.3,4,3.5);
\fill[plateCol, opacity=1, draw=black, thick] 
    ($(plate) + ( 1.8,-2.2,0)$) --
    ($(plate) + ( 1.8,+2.2,0)$) --
    ($(plate) + (-2.2,+2.2,0)$) --
    ($(plate) + (-2.2,-2.2,0)$) -- cycle;
\node at ($ (plate) + (0,-0.2,-0.1) $) [anchor=north] {Device C.G.};
    
\coordinate (imu) at ($ (plate) + (-1.0,1.5,0) $);
\node at ($ (imu) + (0.,0.,0.) $) [anchor=south] {IMU};
\tdplotsetrotatedcoords{20}{80}{0}
\draw [ball color=black,very thin,tdplot_rotated_coords] ( $(imu)$ ) circle (0.1) ;

\draw[thick,->] (imu) -- ++(0, -1, 0) node[anchor=north] {$X_B$};
\draw[thick,->] (imu) -- ++(-1, 0, 0) node[anchor=south west] {$Y_B$};
\draw[thick,->] (imu) -- ++(0,  0,-1) node[anchor=north] {$Z_B$};

\coordinate (cg) at (-1,-1,1.7);
\node at ($ (cg) + (0,1.1,-0.0) $) [anchor=north] {Combined C.G.};

\coordinate (flywheel) at ($ (plate) + (-1.7, -1.6, 0) $);
\draw[thick,->] (flywheel) -- ++(-0.4, 0, 1) node[anchor=west] {$\vec{\omega}_R$};
\node at ($ (flywheel) + (0,0.1,0) $) [anchor=north] {Reaction Wheel};
\tdplotsetrotatedcoords{20}{80}{0}
\draw [ball color=black,very thin,tdplot_rotated_coords] ( $(flywheel)$ ) circle (0.1) ;

\draw [ball color=blue,very thin,tdplot_rotated_coords] ( $(cg)$ ) circle (0.2) ;

\draw[thick,->] (cg) -- ++(-0.5, -1.8, 1.6) node[anchor=east] {$\vec{\omega}$};

\draw [ball color=gray,very thin,tdplot_rotated_coords] ( $(plate)$ ) circle (0.2) ;

\draw [ball color=brown,very thin,tdplot_rotated_coords] ( $(potato)$ ) circle (0.2) ;

\draw[dashed,-{Stealth[length=4mm, width=2.5mm]}] (potato) -- (cg) node[anchor=south] {};
\node at ($ (potato) + (0,0.6,0.2) $) [anchor=south] {$\vec{r}$};
\draw[dashed,-{Stealth[length=4mm, width=2.5mm]}] (plate) -- (cg) node[anchor=south] {};
\node at ($ (plate) - (0,2.5,0.8) $) [anchor=south] {$\vec{s}$};

\end{tikzpicture}
    \caption{``Object'' under test and rigidly attached measuring ``Device''. The body-frame $B$ is defined at the location of the IMU, as shown.}
    \label{fig:diagram}
    \vspace*{-2mm}
\end{figure}

Therefore, it is required to introduce a known torque to the system.
\autoref{fig:diagram} shows a diagram of a possible implementation that attaches a reaction wheel to provide a torque and an \ac{IMU} to measure $\vec{\omega}$.

The following sections show how to model a rigid body with an attached reaction wheel, how the equations of motion during free tumbling can be solved for the inertia, and how to calibrate the possibly unknown inertia of the reaction wheel and the measuring device itself.

\newcommand{\ddt}{\frac{d}{dt}}
\newcommand{\Lfw}{\vec{L_{f}}}
\newcommand{\Lfwd}{\dot{\vec{L_{f}}}}
\newcommand{\Min}{\left(\vec{M}\right)_{\mathrm{in}}}
\newcommand{\Mbody}{\left(\vec{M}\right)_{\mathrm{body}}}

\subsection{Deriving rotational equation of motion}

In an inertial reference frame, Euler's second law states that external moments $\vec{M}$ change the rotational momentum $\vec{L}$ of a rigid-body as $\vec{M} = \left(d\vec{L}/dt \right)_I$,
where the subscript $(\square)_I$ denotes that the derivative is taken with respect to the inertial frame.
The reaction wheel is assumed to be rigidly attached to the body, so that the total angular momentum of the combined body is 
\begin{equation} \label{eq:Lin}
    \vec{L} = I_O\vec{\omega} + I_R\left(\vec{\omega} + \vec{\omega}_R\right),
\end{equation}
where $I_O$ is the body's inertia tensor excluding rotational inertia of the reaction wheel. The distinction between measurement device and object is made at a later stage. $\vec{\omega}$ is the body's angular velocity vector, $I_R$ is the reaction wheel's inertia tensor, and $\vec{\omega}_R$ is the reaction wheel's angular velocity vector with respect to the body frame.

For the subsequent analysis, it is useful to express all inertia tensors in coordinates of the body-fixed frame, around the barycentre, and also to perform the time derivative of \autoref{eq:Lin} with respect to a frame fixed to the rigid-body under test. We denote this with the subscript $(\square)_B$ in the following.
The first equation below is derived in \cite{HerbertGoldstein2002ClassicalMechanics}:
\begin{align}
    \left(\frac{d\vec{L}}{dt}\right)_{I} &= \left(\frac{d\vec{L}}{dt}\right)_{B} + \vec{\omega} \times \vec{L} \\
    \vec{M} &= \left(\dot I_O\vec{\omega} + I_O \dot{\vec{\omega}}
    + \dot I_R\left(\vec{\omega} + \vec{\omega}_R\right) + I_R \left(\dot{\vec{\omega}} + \dot{\vec{\omega}}_R\right)\right)_B  \notag \\
    &\quad+ \vec{\omega} \times \left(I_O\vec{\omega} + I_R\left(\vec{\omega} + \vec{\omega}_R\right)\right)
\end{align}

With respect to the body-fixed frame, $(\dot{I}_O)_B = 0$. Furthermore, if we assume that the reaction wheel attached to the body is axisymmetric around its axis of rotation, then $(\dot{I}_R)_B = 0$ as well since $I_R$ is then independent of its rotation angle.
The final Equation of Motion during free-fall ($\vec{M} = \vec{0}$) is obtained by collecting $\vec{\omega}_R$ terms on the right.
Note that the torque delivered by the motor of the reaction wheel is internal to the combined system of \autoref{eq:Lin}.
Dropping the subscript $B$ for brevity yields:
\begin{equation}
    \underbrace{\left(I_O + I_R\right)}_{I}\dot{\vec{\omega}} + \vec{\omega} \times \underbrace{\left(I_O + I_R\right)}_{I}\vec{\omega} = -I_R\dot{\vec{\omega}}_R - \vec{\omega} \times \left(I_R\vec{\omega}_R\right). \label{eq:eom}
\end{equation}
\vspace*{-10mm}

\subsection{Solving the equation of motion}\label{subsection:solving}
\autoref{eq:eom} is linear in the components of $I$, and we can represent the left-hand side in the form of $\zeta_i\vec{\theta}$, where $\zeta_i$ denotes the transformation matrix associated with the $i$th datapoint, and $\vec{\theta}$ is a vector containing the six unknown components. 
Let us define
\begin{equation}\label{eq: inertia tensor}
    I = \begin{bmatrix}
        I_{xx} & I_{xy} & I_{xz} \\
        I_{xy} & I_{yy} & I_{yz} \\
        I_{xz} & I_{yz} & I_{zz} \\
        \end{bmatrix}%
        \text{ and } \vec{\theta} = \left[\begin{smallmatrix}
            I_{xx}\\
            I_{xy}\\
            I_{yy}\\
            I_{xz}\\
            I_{yz}\\
            I_{zz}
        \end{smallmatrix}\right],
\end{equation}
and rewrite \autoref{eq:eom} as
\begin{equation}
    \underbrace{\zeta_i\vec{\theta}}_{I\dot{\vec{\omega}}_i + \vec{\omega}_i \times \left(I\vec{\omega}_i\right)} 
    = \underbrace{\vec{\mu}_i}_{-I_R\dot{\vec{\omega}}_{R_i} - \vec{\omega}_i \times \left(I_R\vec{\omega}_{R_i}\right)},
    \label{eq:solution_per_datapoint}
\end{equation}
where $\zeta_i$ contains the cross-products of the left-hand side, a known parameter from rigid body dynamics in parameter-linear form \cite{Kawasaki1988}. 

\newcommand{\defeq}{\vcentcolon=}
\newcommand{\eqdef}{=\vcentcolon}

At this point, it is assumed that $I_R$ is known, so that the right-hand-side $\vec{\mu}_i$ can be computed from measurements of the object's rotation rate and the flywheel rotation rate.

\autoref{eq:solution_per_datapoint} is underdetermined, such that multiple observations $i$ are necessary to solve for $\vec\theta$.
Linear regression can then be used with an arbitrary number of data points $i\in{1,\ldots,N}$ by stacking $\zeta_1 \dots \zeta_N$ and $\vec{\mu}_1 \dots \vec{\mu}_N$.

In embedded systems, enough memory to store a long sequence of recorded sensor data may not be available. For on-line applications, \autoref{eq:solution_per_datapoint} can be solved recursively by making use of the recursive least-squares method \cite{Ljung1999}.

\newcommand{\Itest}{(I_{ij})_{\mathrm{comb}}}
\newcommand{\Iobj}{(I_{ij})_{\mathrm{obj}}}
\newcommand{\Iproof}{(I_{ij})_{\mathrm{proof}}}
\newcommand{\Idev}{(I_{ij})_{\mathrm{dev}}}

\newcommand{\Itesth}{(\tilde{I}_{ij})_{\mathrm{comb}}}
\newcommand{\Iobjh}{(\tilde{I}_{ij})_{\mathrm{obj}}}
\newcommand{\Idevh}{(\tilde{I}_{ij})_{\mathrm{dev}}}
\newcommand{\Ir}{I_{R,zz}}

\newcommand{\mtest}{m_{\mathrm{comb}}}
\newcommand{\mobj}{m_{\mathrm{obj}}}
\newcommand{\mproof}{m_{\mathrm{proof}}}
\newcommand{\mdev}{m_{\mathrm{dev}}}

\newcommand{\xtest}{\vec{x}_{\mathrm{comb}}}
\newcommand{\xobj}{\vec{x}_{\mathrm{obj}}}
\newcommand{\xdev}{\vec{x}_{\mathrm{dev}}}

\subsection{Correcting for the inertia of the measurement device}\label{subsection:correction}
The addition of the device to the object has a two-fold effect on the combined mass properties: (1) it moves the centre of mass, and (2) it increases the combined inertia.
When recording sensor data of the combined configuration for use with the solution of the previous section, this combined inertia is computed and needs to be corrected, unless the measurement unit's mass is negligible compared to the object under test.

We introduce subscript $\square_\text{comb}$ to denote properties of the combined configuration, subscript $\square_\text{dev}$ for properties or the measurement device only (assumed known in this section), and $\square_\text{obj}$ for the object-under-test whose properties we ultimately desire.
The Centres of Gravity (CoGs) and the inertia tensors of these 3 configurations are denoted $\vec{x}_\square$ and $\left(I_{ij}\right)_\square$, respectively, and are all expressed in \ac{IMU} body coordinates $B$.
Additionally, we introduce distances $\vec{r} \triangleq \xtest - \xobj$ and $\vec{s} \triangleq \xtest - \xdev$.
Refer again to \autoref{fig:diagram} for a visual representation.

Using these definitions and the general form of the parallel axis theorem,
and $\delta_{ij}$ as the Kronecker delta,
the inertia of the object-under-test is:
\begin{equation}
    \begin{aligned}
        \Iobj =\ &\Itest - \mobj (|\vec{r}|^2 \delta_{ij} - r_i r_j)\ -\\
        & \Idev - \mdev (|\vec{s}|^2 \delta_{ij} - s_i s_j).
    \end{aligned}
    \label{eq:generalisedParallelAxisTheorem}
\end{equation}

As a reminder, we assume the properties of the measurement device $\square_\text{dev}$ are known. $\Itest$ is determined by applying the procedure in the previous section, and the masses can be measured with a scale.
However, $\vec{r}$ and $\vec{s}$ require knowledge of the location of the Centres of Gravity (CoGs) with respect to the \ac{IMU}.

Referring to \autoref{fig:diagram}, we can express $\vec{r}$ as
\begin{align}
    \vec{r} &= \left(\xdev - \xtest\right)\mdev / \mobj.
\end{align}

The crux is now determining $\xtest$, which can be obtained by applying the method shown in \cite{Blaha2024FlyingConfiguration} to the same tumbling data used in the previous section.
The principle can be summarized as follows: in the assumed freefall, the accelerometer sensors in the \ac{IMU} measure the centripetal and tangential acceleration due to the rotation of the body.
Because these angular velocities are known and their derivatives can be estimated, this enables finding the offset between the \ac{IMU} and the centre of gravity within sub-centimetre accuracy.

\subsection{Calibrating reaction wheel inertia and device inertia}\label{subsec:calibration}

The previous sections derived the estimation assuming reaction wheel inertia $I_R$ and measurement device \ac{CoG} $\xdev$ and inertia $\Idev$ are known.
We now propose a calibration procedure that works by applying the method in the previous sections to throw data of 2 configurations with known attached object inertias $\Iobj$.

Similarly to how the left-hand side of \autoref{eq:eom} could be decomposed into $\zeta_i\vec{\theta}$ in \autoref{eq:solution_per_datapoint}, the right-hand side can be expressed as $\eta_i\vec{\varphi}$, where $\vec{\varphi}$ contains the inertia elements of the spinning reaction wheel. $\eta_i$ is a similar cross-product evaluation as shown in \autoref{subsection:solving}. This gives
\begin{equation}
    \zeta_i\vec{\theta} = \eta_i\vec{\varphi},\label{eq:matrix-vec-both}
\end{equation}

where $\vec{\theta}$ and $\vec{\varphi}$ are linearly dependent and cannot be solved simultaneously.
However, if $\vec\omega_R = (0\ 0\ \omega_{R,z})^T$, and $I_{R,xz} = I_{R,yz} = 0$ from axial symmetry, then the right-hand side of \autoref{eq:eom} collapses to $-\Ir\left(\dot{\vec{\omega}}_R + \vec\omega \times \vec\omega_R\right)$.
This allows parametrising $\vec{\varphi} = ( 0\ 0\ 0\ 0\ 0\ \Ir )^T$.

We first set $\Ir = 1$, and perform the data analysis procedure outlined in \autoref{subsection:solving} for throws of 2 known $\Iobj$ and $\mobj$.
These analyses yield the unitless $\Itesth$, which needs to be scaled by the true $\Ir$ (see \autoref{eq:matrix-vec-both}) to yield the true combined inertias
\begin{align}
\begin{split}
    \Itest^1 &= \Ir\ \Itesth^1\\
    \Itest^2 &= \Ir\ \Itesth^2 .
\end{split}
\end{align}

Plugging this into 2 instances of \autoref{eq:generalisedParallelAxisTheorem} gives 12 equations in 10 unknowns ($\Ir$, $\Idev$, $\xdev$). Solving the system can be simplified greatly by choosing that one of the throws be performed without any body attached, which we call the device-only configuration with $\Iobj^1 = 0$ and $\mobj^1 = 0$.

The second configuration is called the proof-body configuration with known and non-zero $\Iproof \triangleq \Iobj^2$ and $\mproof \triangleq \mobj^2$.
The 2 instances of \autoref{eq:generalisedParallelAxisTheorem} collapse to
\begin{align}
    \Ir\ \Itesth^1 =\ &\Idev \\
    \begin{split}
        \Ir\ \Itesth^2 =\ &\Iproof + \mproof (|\vec{r}|^2 \delta_{ij} - r_ir_j)\ +\\ 
        \ &\Idev + \mdev (|\vec{s}|^2  \delta_{ij} - s_is_j),
        \label{eq:j_outside}
    \end{split}
\end{align}
which can be trivially reduced to 6 equations in 1 unknown ($\Ir$) by substituting the first into the second:
\begin{align}
\begin{split}
    \Ir&\left(\Itesth^2 - \Itesth^1\right) \\
       & = \Iproof + \mproof (|\vec{r}|^2 \delta_{ij} - r_i r_j) \\
    &\quad\quad\quad\quad\quad\ + \mdev (|\vec{s}|^2  \delta_{ij} - s_i s_j)
\end{split}
\end{align}

The right-hand side is known, and the left-hand side is known up to the scalar $\Ir$.
By transforming the two sides into vectors following \autoref{eq: inertia tensor}, we can finally write the least-square solution for $\Ir$ as the orthogonal projection of the right-hand side onto the left-hand side:
\begin{equation}
    \Ir = (\vec{\theta}_{\text{lhs}}^T \vec{\theta}_{\text{rhs}}) / |\vec{\theta}_{\text{lhs}}|^2.
\end{equation}

\begin{figure}
    \centering
    \includegraphics[trim={0.5cm, 0.5cm, 0.5cm, 0.5cm},clip,width=0.95\linewidth]{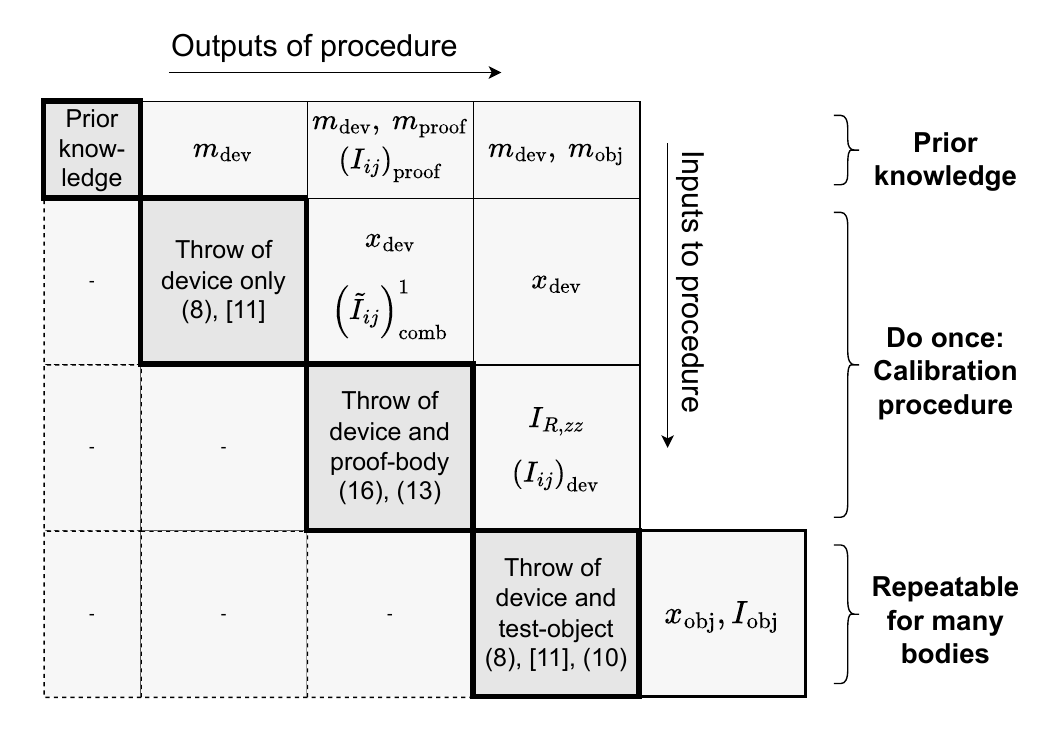}
    \caption{Depiction of the method. Subprocedures on the diagonal, required inputs vertically above, outputs on the right.}
    \label{fig:n2chart}
    \vspace*{-4mm}
\end{figure}

\autoref{fig:n2chart} visually summarizes the procedures.

\subsection{Error metrics}

In order to distinguish between the error in the measured inertia magnitudes and the error in the estimation of their rotational alignment, we propose two error metrics.

First, we define a frame-independent relative inertial magnitude error, by comparing the principal moments of inertia. Specifically, we use the ratio of the estimation error of the principal moments of inertia $\Delta \lambda_1$, $\Delta \lambda_2$, $\Delta \lambda_3$ to the true principal moments of inertia $\lambda_1$, $\lambda_2$, $\lambda_3$:
\begin{equation}
    \varepsilon = \sqrt{\frac{(\Delta\lambda_1)^2 + (\Delta\lambda_2)^2 + (\Delta\lambda_3)^2}{\lambda_1^2 + \lambda_2^2 + \lambda_3^2}}.
\end{equation}

The spatial transformation from the matrix representation of $I_{\mathrm{obj}}$ to the diagonalised representation is a rotation, which can be found using the singular value decomposition (SVD) of $I = U\Sigma V^T$. We define the alignment error $\psi$ as the geodesic distance between the ground truth $\tilde{U}$ and the estimated $U$:
\begin{equation}
    \psi = \arccos\Big[0.5\ \mathrm{Tr}(\tilde{U}^{T} U) - 0.5\Big].
\end{equation}

\section{Verification and validation}\label{sec:experiments}
This section describes simulations and experiments performed to validate the procedure developed previously.

\begin{figure}[bt]
    \centering
    \begin{subfigure}[b]{0.48\linewidth}
        \centering
        \includegraphics[trim={0cm, 0cm, 0cm, 0cm},clip,width=0.9\columnwidth]{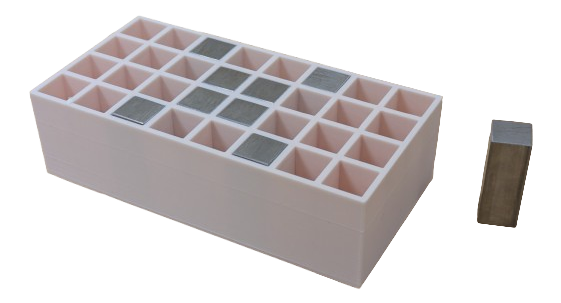}
        \caption{"Grid body" and weights.}
        \label{fig:gridmass}
    \end{subfigure}%
    \hfill%
    \begin{subfigure}[b]{0.48\linewidth}
        \centering
        \tikzset{
          font={\fontsize{7pt}{12}\selectfont}}
        \tdplotsetmaincoords{45}{140}
        \begin{tikzpicture}[tdplot_main_coords, scale=0.7]
            \node[anchor=south west,inner sep=0] at (0,0) {\includegraphics[trim={7cm 3cm 14cm 4cm},clip,width=0.9\columnwidth]{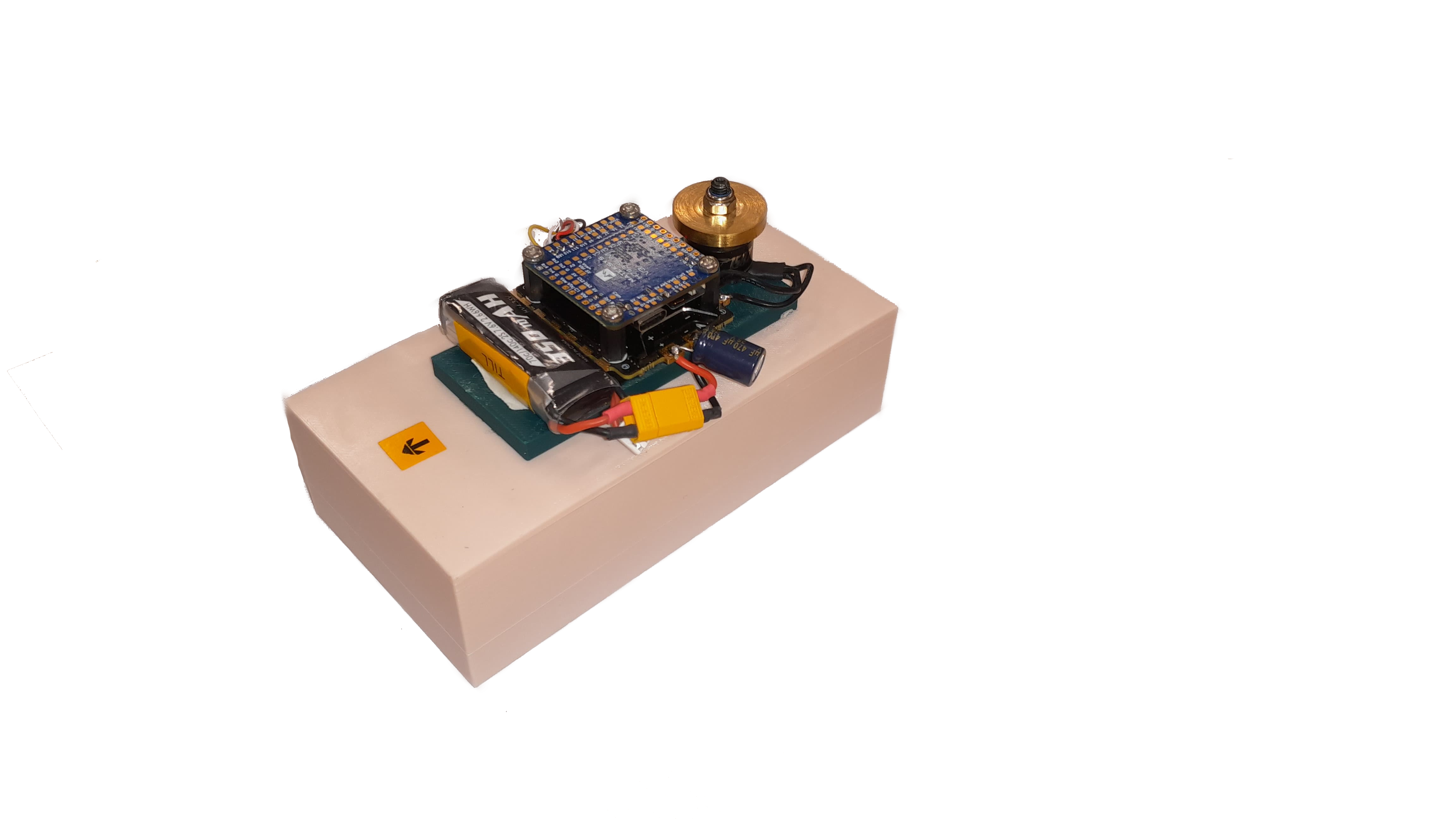}};
            
            \tdplotsetrotatedcoords{50}{80}{0}
            
            \coordinate (imu) at ($ (-5.7,-1.6,0) $);
            \draw[thick,->,color=white] (imu) -- ++(0, -0.8, 0) node[anchor=north] {$X_B$};
            \draw[thick,->,color=white] (imu) -- ++(-0.8, 0, 0) node[anchor=north] {$Y_B$};
            \draw[thick,->,color=white] (imu) -- ++(0,  0,-0.8) node[anchor=east] {$Z_B$};
            \draw [ball color=white,very thin,tdplot_rotated_coords] ( $(imu)$ ) circle (0.125) ;
            \node at ($ (imu) + (0.2,0.,0.1) $) [anchor=north west, color=white] {{ }IMU};
            
            \coordinate (cg) at ($ (imu) + (-0.5, -0.8, -2.35) $);
            \draw [ball color=blue,very thin,tdplot_rotated_coords] ( $(cg)$ ) circle (0.125) ;
            \node at ($ (cg) $) [anchor=west, color=white] {{ }Comb. C.G.};

            \coordinate (potato) at ($ (imu) + (-0.45, -0.8, -3.0) $);
            \draw [ball color=brown,very thin,tdplot_rotated_coords] ( $(potato)$ ) circle (0.125) ;
            \node at ($ (potato) $) [anchor=north east, color=white] {Object C.G.};
        \end{tikzpicture}
        \caption{Full setup. See \autoref{fig:diagram}.}
        \label{fig:validation_setup}
    \end{subfigure}
    \caption{Overview of the experimental hardware.}\label{fig:hardware}
    \vspace*{-2mm}
\end{figure}
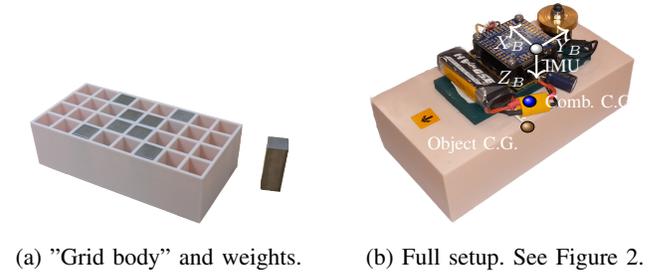

\subsection{Experimental Hardware}

We collected the required data using a consumer-grade multirotor flight controller with onboard MEMS-IMU (TDK InvenSense\texttrademark{} ICM-42688-P).
The reaction wheel is actuated using a brushless DC motor, controlled by an \ac{ESC} with motor rotation speed feedback.
The device
weighs \num{100}~\si{\gram}, and is pictured in \autoref{fig:validation_setup}.
The sampling frequency was set to \num{4}~\si{\kilo\Hz}.

The proof body required for the calibration procedure was chosen to be a cuboid of homogenous aluminium (not shown), machined to \num{70}$\times$\num{60}$\times$\num{30}~\si{\mm}, so that its inertia is accurately determined from geometry and mass, and significantly larger than those of the test device.

To demonstrate that the method is accurate for a versatile range of inertial properties, a grid-frame was 3D printed with pre-shaped holes for steel cuboids (shown in \autoref{fig:gridmass}).
This allows for a large number of configurations of weights, providing a large number of possible inertia tensors.
Also, for this body, we assume that we can calculate its ground-truth inertia from the geometry and mass of a given configuration.

\subsection{Experimental procedure}

We performed the calibration procedure from \autoref{subsec:calibration} by combining data from multiple device-only and device + proof-body throws. With the resulting $\xdev$, $\Idev$, and reaction wheel $\Ir$, 4 different configurations of the grid body were thrown 10 or more times each. 
The method from \autoref{subsection:solving} and \autoref{subsection:correction} was run on the data of each throw \emph{individually}, and then the errors were aggregated.

The arrangement and coordinate system used in the validation tests are shown in \autoref{fig:validation_setup}. The reaction wheel motor is programmed to deliver a pulse, shown in \autoref{fig:simulation}.

High-frequency vibration and intrinsic \ac{IMU} noise was removed by applying a zero-phase 4th-order Butterworth low-pass filter with a \num{20}~\si{\Hz} cutoff to all signals.

\subsection{Simulations}

\begin{figure}[bt]
    \centering
    \includegraphics[width=\linewidth]{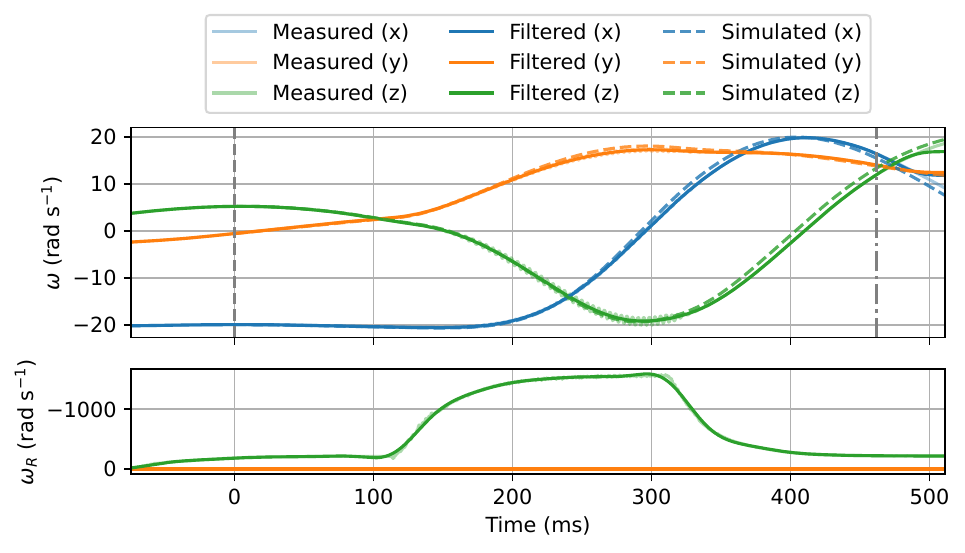}
    \caption{Simulation of the angular velocities, based on an initial condition at the dashed line, and the filtered flywheel rotation compared to measured and filtered angular velocities.}
    \label{fig:simulation}
    \vspace*{-4mm}
\end{figure}
First, to demonstrate the validity of \autoref{eq:eom}, several simulations were performed, based on a single data point in the falling phase, and the measured inertia tensor.
For this, the measured reaction wheel angular rotation was filtered and then used to evaluate the right-hand side. A forward Euler integration scheme was then used to evaluate how well the measured inertia tensor produces the same rotation, given the same (filtered) input. One such simulation can be seen in \autoref{fig:simulation}, where the initial condition is at $t=0$.

Second, we used Monte Carlo simulations to verify that solving \autoref{eq:solution_per_datapoint} with the ordinary least-squares method is robust to measurement noise, initial condition, and body shape.

For all simulations, we sample 3 principal moments of inertia so that $\operatorname{Tr}(I) =\ $\si{2000}~\si{\kilogram\milli\metre\squared} and the condition number $\kappa(I)$ (serving as a measure of elongation of the body) is uniformly distributed on $[1,\ 5]$.
To show invariance to the axis systems, we randomly transform these principal inertias on $SO(3)$, and choose a random direction for the initial body rotation $\omega_0$. Furthermore, $\lvert \omega_0 \rvert_2 \in [2\pi,\ 6\pi]$~\si{\radian\per\second}.

If two principal inertias are close to equal, the principal axis system is not unique. Therefore, we evaluate for each sampled body the minimum relative distance of the principal inertias:
\begin{equation}
    \min\Delta\bar\sigma \triangleq \min_{i\neq j}\frac{\mid I_i - I_j \mid}{\max(I_i, I_j)}.
\end{equation}

In most simulations, we use the reaction wheel impulse $L_R$ and measurement noise densities ($N_\omega$, $N_{\omega_R}$) observed in reality, but we also provide simulations where these are varied.

\section{Results}\label{sec:results}
\setlength{\tabcolsep}{5pt} 
\begin{table*}
\vspace*{2mm}
\centering
\caption{Aggregated experiment results. Percentage inertia component errors $\varepsilon$ and alignment error $\Psi$ are shown for the different configurations, as well as \ac{CoG} estimates (value along $X_B,\ Y_B$ unvalidated, as no ground truth exists).}
\label{tab:grid_results}
\begin{tabular}{crrr|r|rrrr|rr}
\multicolumn{4}{c|}{\bfseries Ground truth from geometry} & & \multicolumn{6}{c}{\bfseries Our algorithms, run on data of single throws} \\
\hline
Configuration & m [\si{\kg}] & CoGz [\si{mm}] & diag$I$ [\si{\kg\square\mm}] & N & mean$(\varepsilon)$ & max$(\varepsilon)$ & mean$(\Psi)$ & max$(\Psi)$ & mean$(\hat{x})$ [\si{mm}] & stddev$(\hat{x})$ [\si{mm}] \\
\hline
E: {\makeEmptyGrid} &
    0.178 & 43.4 & [368 123 431] & 10 & 1.6\% & 2.3\% & 2.1\si{\degree} & 2.4\si{\degree} & [10.7  1.8 43.1] & [0.2  0.23 0.24] \\
\hline
A: \makeGrid{1/2, 1/3, 8/2, 8/3} &
    0.459 & 45.3 & [1525  190 1577] & 11 & 1.7\% & 6.6\% & 3.5\si{\degree} & 5.5\si{\degree} & [10.7  2.3 45.9] & [0.18 0.46 0.49] \\
\hline
B: \makeGrid{3/1, 3/4, 4/2, 4/3, 5/2, 5/3, 6/1, 6/4} &
    0.739 & 45.8 & [682 437 906] & 12 & 1.8\% & 4.3\% & 2.1\si{\degree} & 2.2\si{\degree} & [10.7  1.4 45.6] & [0.14 0.06 0.09] \\
\hline
C: \makeGrid{1/1, 1/4, 2/2, 2/3, 3/1, 3/4, 4/2, 4/3, 5/2, 5/3, 6/1, 6/4, 7/2, 7/3, 8/1, 8/4} &
    1.300 & 46.1 & [2448 750 2835] & 12 & 2.5\% & 4.1\% & 1.6\si{\degree} & 1.9\si{\degree} & [10.5  1.7 46.1] & [0.06 0.28 0.19] \\
\end{tabular}
\vspace*{-2mm}
\end{table*}

\autoref{tab:grid_results} shows the aggregated experiment results with different configurations of the grid body. The inertia of the device and the reaction wheel have been calibrated from throws with and without the aluminium proof block attached.

In the average case, the errors after processing data from single throws are below $3\%$ of the magnitude of the true moments of inertia, and below $4$\si{\degree} for the alignment of the principal axis system.
For any configuration, the maximum observed error from single throw data is $6.6\%$, and $5.5$\si\degree, which both occur at configuration A.

The computation of the location of the \ac{CoG} of the object (not the combination object + device!) from single throw data is also shown in the table. It was repeatable up to a standard deviation of below \num{0.5}~\si{mm}; this worst-case also occurs at configuration A.
However, no ground truth is available for $x$ and $y$ of this quantity, as the precise location of the measurement device on the grid body has not been recorded.
The height of the IMU with respect to the body was recorded, and so the $z$-distance between the calculated \ac{CoG} and \ac{IMU} is shown in the CoGz column of \autoref{tab:grid_results}.
It is within \num{0.6}~\si{mm} of the estimated $z$-coordinate, worst case.

\begin{figure*}[t]
    \centering
    \begin{subfigure}[b]{0.33\linewidth}
        \centering
        \includegraphics[trim={0.4cm, 0.3cm, 10cm, 6cm},clip,scale=0.75]{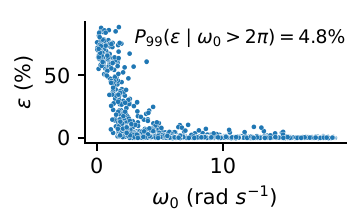}
        \caption{Initial rotation magnitude (axis random)}
        \label{fig:initial_condition}
    \end{subfigure}%
    \begin{subfigure}[b]{0.66\linewidth}
        \centering
        \includegraphics[trim={0.4cm, 0.3cm, 10.1cm, 6cm},clip,scale=0.75]{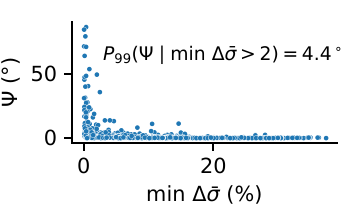}
        \includegraphics[trim={0.4cm, 0.3cm, 10cm, 6cm},clip,scale=0.75]{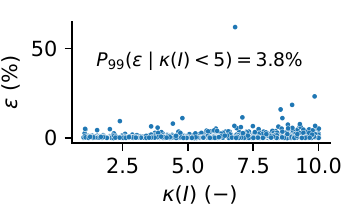}
        \caption{Object shape: principal inertia similarity, and elongation.}
        \label{fig:body_type}
    \end{subfigure}
    \begin{subfigure}[b]{0.24\linewidth}
        \centering
        \includegraphics[trim={1cm, 1.2cm, 20.5cm, 11.6cm},clip,scale=0.75]{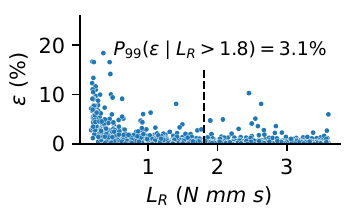}
        \caption{Reaction wheel impulse}
        \label{fig:reactionwheel}
    \end{subfigure}%
    \begin{subfigure}[b]{0.24\linewidth}
        \centering
        \includegraphics[trim={7cm, 1.2cm, 15.5cm, 11.6cm},clip,scale=0.75]{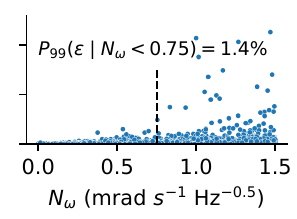}
        \caption{Gyroscope noise}
        \label{fig:gyronoise}
    \end{subfigure}%
    \begin{subfigure}[b]{0.24\linewidth}
        \centering
        \includegraphics[trim={12.3cm, 1.2cm, 10.5cm, 11.6cm},clip,scale=0.75]{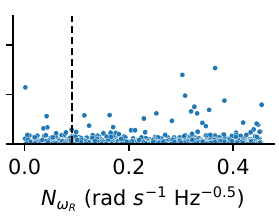}
        \caption{Reac. wheel speed noise}
        \label{fig:reactionwheelnoise}
    \end{subfigure}%
    \begin{subfigure}[b]{0.24\linewidth}
        \centering
        \includegraphics[trim={17.4cm, 1.2cm, 5.3cm, 11.6cm},clip,scale=0.75]{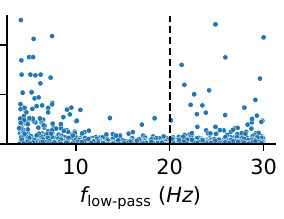}
        \caption{Low pass filter}
        \label{fig:lpcutoff}
    \end{subfigure}%
    \caption{6 Monte Carlo simulations (N=2000 each), showing the influence of initial condition, body shape, reaction wheel size and filtering. A handful of outliers at the edges of the parameter ranges are not shown on (c)-(f) for better scaling.}\label{fig:monte_carlo}
    \vspace*{-3mm}
\end{figure*}

\autoref{fig:monte_carlo} shows the results of the Monte Carlo simulations, with empirical 99th percentiles of the accuracy metrics. Vertical bars indicate the conditions and choices of the experiments.

\section{Discussion} \label{sec:discussion}
The simulation results are in good agreement with the accuracy observed in the experiments, allowing us to use the simulation results to understand the sensitivities and limitations of our method.

The identification is neither sensitive to the relative alignment of sensor and principal axes, nor to the initial axis of rotation; both were randomized in the simulations.
However, it is important to throw with spin in excess of 1 revolution per second, as shown in \autoref{fig:initial_condition}.

The estimation is also robust to the shape of the body as long as the largest principal moment of inertia is not more than 5 times that of the smallest axis, and any two principal moments are more than 2\% different. This limitation is illustrated in \autoref{fig:body_type} and explains the decreased accuracy of configuration A in \autoref{tab:grid_results}.
In these cases, the measurement device can be placed to create a combined body that has its principal inertias more widely spaced, improving accuracy.

After ensuring enough spin, the single biggest factor determining accuracy is the noise level present in the sensors (\ref{fig:gyronoise}).
We observed around \num{0.75}~\si{(\milli\radian\per\second)\per\sqrt\Hz} for the gyroscope due to vibrations caused by the reaction wheel.
The gyroscope data sheet indicates only \num{0.05}~\si{(\milli\radian\per\second)\per\sqrt\Hz}, which highlights the need for a well-balanced reaction wheel.

From the simulations, it follows that filtering inputs with a cutoff between \num{10} and \num{20}~\si{\Hz} is optimal for the observed noise level. Too low, and relevant frequencies are dampened; too high, and the signal-to-noise ratio becomes too low.
Low-frequency vibrations from violating the rigid-body assumption (e.g., from loose wires) can have detrimental effects on the accuracy and must be avoided.

The biases of the consumer-grade \ac{IMU} are typically around \num{\pm9}~\si{\milli\radian\per\second} and \num{\pm0.02}~\si{g} respectively, as per datasheet (TDK InvenSense\texttrademark{} ICM-42688-P); less than $1\%$ of the actually occurring values during the throws. 
Both the gyroscope and accelerometer are also specified with $0.5\%$ scaling factor offset and $0.1\%$ scale nonlinearity over the measuring range.
It is expected that these effects explain part of the estimation error. 
Since the biases are not expected to change during the short throw, they could be corrected for by measuring the zero-output at rest on a level surface just before the throw.

The impulse delivered by the reaction wheel should be sufficiently large in order to influence the motion well above the noise floor. Simulation (\ref{fig:reactionwheel}) reveals that our choice of \num{1.8}~\si{\newton\milli\metre\second} for bodies in the \num{1000}~\si{\kilogram\milli\metre\squared} range is sufficient, but it should not be less than this ratio.
At the same time, when the reaction wheel impulse is too high, this risks exceeding the sensing limits of the gyroscope.
Vibrations originating from an imbalanced reaction wheel then lead to noise, can saturate the accelerometer, and reduce accuracy. 
With these conditions met, the accuracy in our experiments is comparable to pendulum methods (reported 2.5 to 5\% \cite{Dowling2006TheInertia} or 1.2 to 5.4\% \cite{setati2022}).

The presented model does not account for the air resistance acting on the rotating body. Therefore, the method works best on densely packed bodies that are ideally roughly spherical. This limitation is not easily quantified and therefore, left qualitative.

\section{Conclusions}
We have shown a method to estimate the Centre of Gravity (CoG) and mass moments of inertia of a rigid body from data recorded on a small attached measurement device during a manual throw.
The device comprises a 3-axis gyroscope and accelerometer, as well as a small reaction wheel whose angular impulse relative to body inertia should be greater than \num{1.8}~\si{\newton\metre\second\per(\kilogram\metre\squared)}. A dynamical model of the system and data from free-spinning can be used to fit the inertial properties.

Calibration of the device and reaction wheel inertia was effectively achieved from throws of the device with and without an attached body of known mass properties. 
Then, the device can be attached to other rigid bodies, and the corresponding inertia matrix and \ac{CoG} locations can be estimated, requiring only the mass of the body to be known.
The mean errors in the magnitude of the principal inertias are around 2\% from a single throw.
The directions of the principal axes can be computed on average within about 2.5$^\circ$.
The inertial estimation from gyroscope data is invariant of initial condition and object rotation. However, restrictions on object elongation and measurement noise have been found.

If vertical spinning throws are convenient, and aerodynamic effects are not a factor, then the accuracy is comparable to that expected of pendulum methods, but requires less setup and directly provides \ac{CoG} location and the full inertia tensor.

\section*{Acknowledgement}

ChatGPT provided assistance in writing the analysis code.




\end{document}